\documentclass{bmvc2k}
\usepackage{multirow}
\usepackage{booktabs} 
\usepackage{hyperref}
\usepackage{graphicx}
\usepackage{enumitem}  
\usepackage{amsmath}

\title{Class-Weighted Convolutional Features for Visual Instance Search}

\addauthor{Albert Jimenez}{albertjimenezsanfiz@gmail.com}{12}
\addauthor{Jose M. Alvarez}{http://www.josemalvarez.net}{2}
\addauthor{Xavier Giro-i-Nieto}{xavier.giro@upc.edu}{1}

\addinstitution{
 Image Processing Group\\
 Universitat Politecnica de Catalunya\\
 Barcelona, Spain
}
\addinstitution{
CSIRO / Data61\\
Canberra, Australia
}

\runninghead{Jimenez, Alvarez, Giro-i-Nieto}{Class-Weighted Conv. Feats}

\begin{document}

\maketitle

\begin{abstract}
Image retrieval in realistic scenarios targets large dynamic datasets of unlabeled images. In these cases, training or fine-tuning a model every time new images are added to the database is neither efficient nor scalable. Convolutional neural networks trained for image classification over large datasets have been proven effective feature extractors for image retrieval. The most successful approaches are based on encoding the activations of convolutional layers, as they convey the image spatial information. In this paper, we go beyond this spatial information and propose a local-aware encoding of convolutional features based on semantic information predicted in the target image. To this end, we obtain the most discriminative regions of an image using Class Activation Maps (CAMs). CAMs are based on the knowledge contained in the network and therefore, our approach, has the additional advantage of not requiring external information. In addition, we use CAMs to generate object proposals during an unsupervised re-ranking stage after a first fast search. Our experiments on two public available datasets for instance retrieval, Oxford5k and Paris6k, demonstrate the competitiveness of our approach outperforming the current state-of-the-art when using off-the-shelf models trained on ImageNet. Our code is publicly available at \url{http://imatge-upc.github.io/retrieval-2017-cam/}.
\end{abstract}

\section{Introduction}
\label{sec:intro}
Content-based Image Retrieval (CBIR) and, in particular, object retrieval (instance search) is a very active field in computer vision. Given an image containing the object of interest (visual query), a search engine is expected to explore a large dataset to build a ranked list of images depicting the query object. This task has been addressed in multiple ways: from learning efficient representations~\cite{perronnin2010large, radenovic2015multiple} and smart codebooks~\cite{philbin2007object, avrithis2012approximate}, to refining a first set of quick and approximate results with query expansion~\cite{chum2011total,tolias2014visual,iscen2016efficient} or spatial verification~\cite{philbin2007object, shen2014spatially}.

Convolutional neural networks trained on large scale datasets have the ability of transferring the learned knowledge from one dataset to another~\cite{yosinski2014transferable}. This property is specially important for the image retrieval problem, where the classic study case targets a large and growing dataset of unlabeled images. Therefore, approaches where a CNN is re-trained every time new images are added does not scale well in a practical situation. 

Many works in the literature focus on using a pre-trained CNN as feature extractor and, in some cases, enhancing these features by performing a fine-tuning step on a custom dataset. For instance,~\cite{babenko2014neural} and~\cite{gong2014multi} use the activations of the fully-connected layers while more recent works have demonstrated that the activations of convolutional layers convey the spatial information and thus, provide better performance for object retrieval~\cite{babenko2015aggregating}. Following this observation, several works have based their approach on combining convolutional features with regions of interest inside the image~\cite{razavian2014baseline, babenko2015aggregating, tolias2015particular, kalantidis2015cross}. More recent works have focused on applying supervised learning to fine-tune CNNs using a similarity oriented loss such as ranking~\cite{gordo2016end} or pairwise similarity~\cite{radenovic2016cnn} to adapt the CNN and boost the performance of the resulting representations. However, this fine-tuning step has the main drawback of having to spend large efforts on collecting, annotating and cleaning a large dataset, which sometimes is not feasible.

In this paper, we aim at encoding images into compact representations taking into account the semantics of the image and using only the knowledge built in the network. Semantic information has been considered before in the context of image retrieval. For instance,~\cite{zhang2013semantic} proposed a method to combine semantic attributes and local features to compute inverted indexes for fast retrieval. Similarly, in \cite{felix2012weak}, the authors use an embedding of weak semantic attributes. However, most of these methods do not associate image regions with the objects in the image, as this process usually relies in other expensive approaches like object detectors. Here, by contrast, we use convolutional features weighted by a soft attention model over the classes contained in the image. The key idea of our approach is exploiting the transferability of the information encoded in a CNN, not only in its features, but also in its ability to focus the attention on the most representative regions of the image. To this end, we use Class Activation Maps (CAMs)~\cite{zhou2015cnnlocalization} to generate semantic-aware weights for convolutional features extracted from the convolutional layers of a network.

The main contributions of this paper are: First, we propose to encode images based on their semantic information by using CAMs to spatially weight convolutional features. Second, we propose to use the object mappings given by CAMs to compute fast regions of interest for a posterior re-ranking stage. Finally, we set a new state-off-the art in Oxford5k and Paris6k using off-the-shelf features.


\section{Related Work}\label{sec:related_work}
Following the success of CNNs for the task of image classification, recent retrieval works have replaced hand-crafted features for representations obtained from off-the-shelf CNNs. For instance, in~\cite{babenko2014neural}, the authors use features extracted from the fully-connected layers of the networks. An extension to local analysis was presented in~\cite{sharif2014cnn}, where features were extracted over a fixed set of regions at different scales defined over the image.

Later, it was observed that features from convolutional layers convey the spatial information of images making them more useful for the task of retrieval. Based on this observation, recent approaches focus on combining convolutional features with different methods to estimate areas of interest within the image. For instance, R-MAC~\cite{tolias2015particular} and BoW~\cite{mohedano2016bags} use a fixed grid of regions,~\cite{sharif2014cnn} considers random regions, and SPoC~\cite{babenko2015aggregating} assumes that the relevant content is in the center of the image (dataset bias). These approaches show how focusing on local regions of the image improves performance. However, the computation of these regions is based on heuristics and randomness. By contrast, in this paper we focus on obtaining local regions based on image contents.

\begin{figure}[!t]
\centering
	\includegraphics[width=\textwidth]{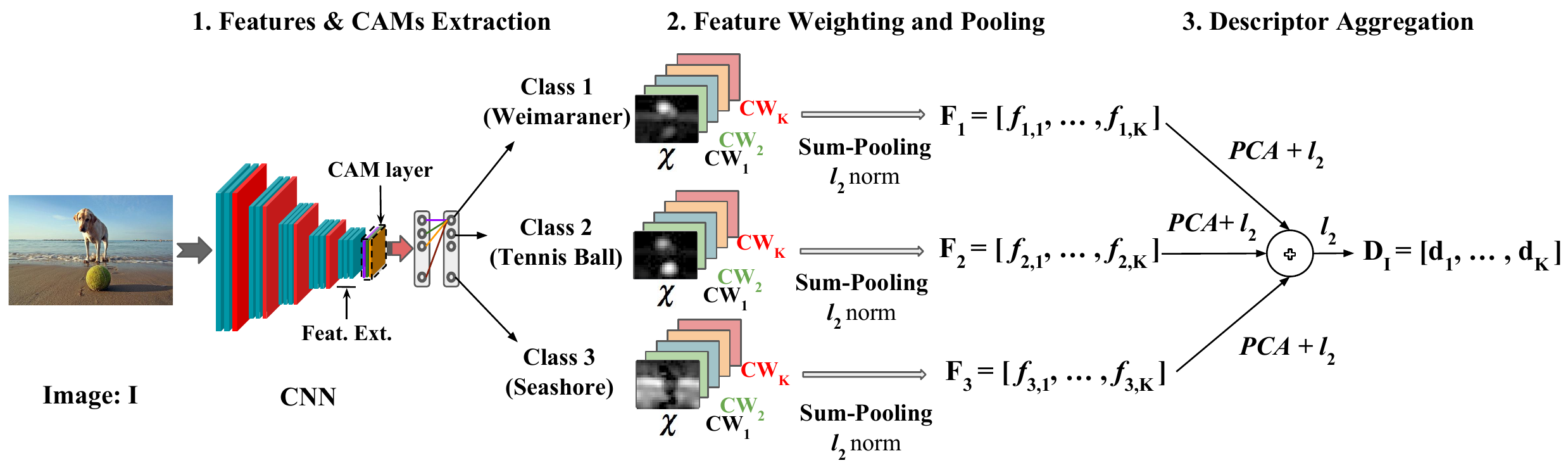}
    \vspace{-0.70cm}
    \caption{Our image encoding pipeline. An image $I$ is used as input to a convolutional neural network. After a forward pass, feature maps are spatially weighted by each CAM, sum-pooled and then, weighted by channel as detailed in Eq.~\ref{eq:channelweight}. Then, normalized class vectors $F_c$ are generated by applying PCA and \textit{l\textsubscript{2}} normalization. Finally, these class vectors are aggregated and normalized to build the final compact image representation $D_I$.}
    \label{fig:descriptor_pipeline}
    \vspace{-0.4cm}
\end{figure}

In this work, we aim at extracting features with focus on local areas that depend on the contents of the image, as other authors have explored in the past. For instance, in~\cite{gordo2016end, salvador2016faster}, a region proposal network is trained for each query object. However, this solution does not scale well as it is a computational intensive process that must be run at query time, both for the training, and for the analysis of a large scale dataset at search time. Other approaches use an additional model to predict regions of interest for each image. For example, the work in~\cite{reyes2016my} uses saliency maps generated by an eye gaze predictor to weight the convolutional features. However, this option requires additional computation of the saliency maps and therefore duplicates the computational effort of indexing the database. Yet another approach is proposed by the CroW model~\cite{kalantidis2015cross}. This model estimates a spatial weighting of the features as a combination of convolutional feature maps across all channels of the layer. As a result, features at locations with salient visual content are boosted while weights in non-salient locations are decreased. This weighting scheme can be efficiently computed in a single forward pass. However, it does not explicitly leverage semantic information contained in the model. In the next section, we present our approach based on Class Activation Maps~\cite{zhou2015cnnlocalization} to exploit the predicted classes and obtain semantic-aware spatial weights for convolutional features.

\section{Class-Weighted Convolutional Features}\label{sec:proposal}
\begin{figure}[!t]
	\centering
	\hspace{-0.05cm}\begin{tabular}{ccccc}
    \hspace{-0.15cm}\scriptsize{Input Image}&\hspace{-0.35cm}\scriptsize{VGG-16-CAM}&\hspace{-0.25cm}\scriptsize{DecomposeMe}&\hspace{-0.35cm}\scriptsize{ResNet-50}&\hspace{-0.35cm}\scriptsize{DenseNet-161}\\
	\hspace{-0.15cm}\includegraphics[width=0.195\textwidth]{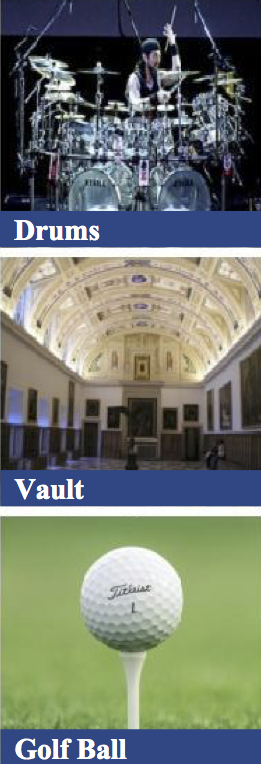}&
    \hspace{-0.35cm}\includegraphics[width=0.195\textwidth]{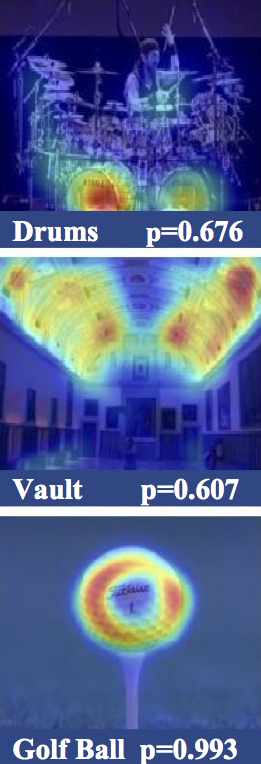}&
    \hspace{-0.35cm}\includegraphics[width=0.195\textwidth]{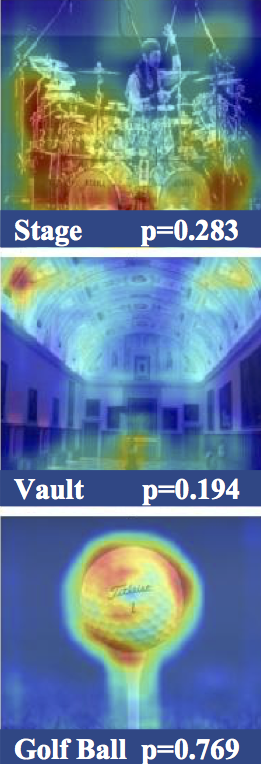}&
	\hspace{-0.35cm}\includegraphics[width=0.195\textwidth]{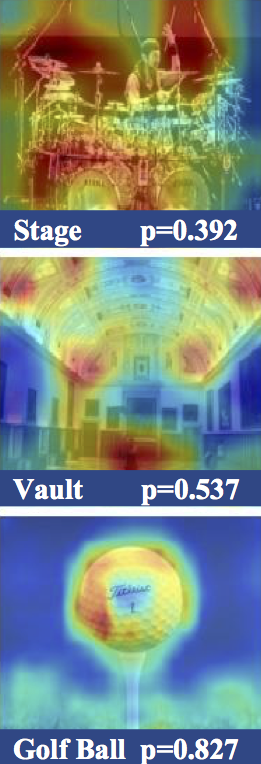}&
    \hspace{-0.35cm}\includegraphics[width=0.195\textwidth]{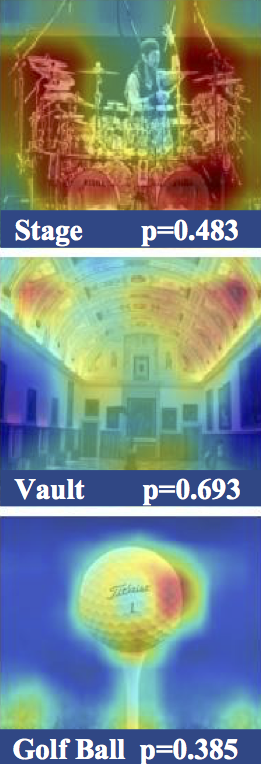}\\        
    \end{tabular}
    \centering
    \caption{{Qualitative CAMs obtained using several network architectures such as VGG-16-CAM~\cite{zhou2015cnnlocalization}, DecomposeMe~\cite{alvarez2016decomposeme}, ResNet-50~\cite{he2016deep} or DenseNet-161~\cite{huang2016densely}. Each example shows the top predicted class and the probability assigned by the model to that class. For the input image, we show the ground truth assigned to that particular image.}}\label{fig:qualitative_cams}
\vspace{-0.25cm}
\end{figure}

In this section, we first review Class Activation Maps and then outline our proposed pipeline for encoding images into compact representations.
\subsection{Class Activation Maps} 
\label{subs:cams}
Class Activation Maps (CAMs)~\cite{zhou2015cnnlocalization} were proposed as a method to estimate relevant pixels of the image that were most attended by the CNN when predicting each class.
The computation of CAMs is a straightforward process in most state-of-the-art CNN architectures for image classification. In short, the last fully-connected layers are replaced with a Global Average Pooling (GAP) layer and a linear classifier. Optionally, an additional convolutional layer can be added before the GAP (\textit{CAM layer}) to recover the accuracy drop after removing the fully-connected layers. In architectures where the layer before the classifier is a GAP layer, CAMs can be directly extracted without any modification.

Given an output class \textit{c}, its CAM is computed as a linear combination of the feature maps in the last convolutional layer, weighted by the class weights learned by the linear classifier. More precisely, the computation of the CAM for the $c$-th class is as follows:
\vspace{-0.1cm}
\begin{equation}
\label{eq:CAMs}
CAM_{c} = \sum_{k=1}^K conv_{k}\cdot w_{k, c},
\end{equation}
where $conv_k$ is the $k$-th feature map of the convolutional layer before the GAP layer, and $w_{k, c}$ is the weight associated with the $k$-th feature map and the $c$-th class. Notice that, as we are applying a global average pooling before the classifier, the CAM architecture does not depend on the input image size. 

Given a CAM it is possible to extract bounding boxes to estimate the localization of objects~\cite{zhou2015cnnlocalization}. The process consists of setting a threshold based on the normalized intensity of the CAM heat map values and then set to zero all values below that threshold. The region of of interest is defined as the bounding box that covers the largest connected element. 
\subsection{Image Encoding Pipeline} \label{ss:encoding}
The image encoding pipeline is depicted in Figure~\ref{fig:descriptor_pipeline} and consists of three main stages: Features and CAM extraction, feature weighting and pooling and descriptor aggregation.

\textbf{Features and CAMs Extraction:} Input images are feed-forwarded through the CNN to compute, in a single pass, convolutional features of the selected layer with \textit{K} feature maps ($\chi$) with a resolution of $W\times H$. In the same forward pass, we also compute CAMs to highlight the class-specific discriminative regions attended by the network. These CAMs are normalized to fall in the range $\left[0, 1\right]$ and resized to match the resolution of the selected convolutional feature maps.

\begin{figure}[!t]
	\centering	
	\hspace{-0.15cm}\includegraphics[width=0.99\textwidth]{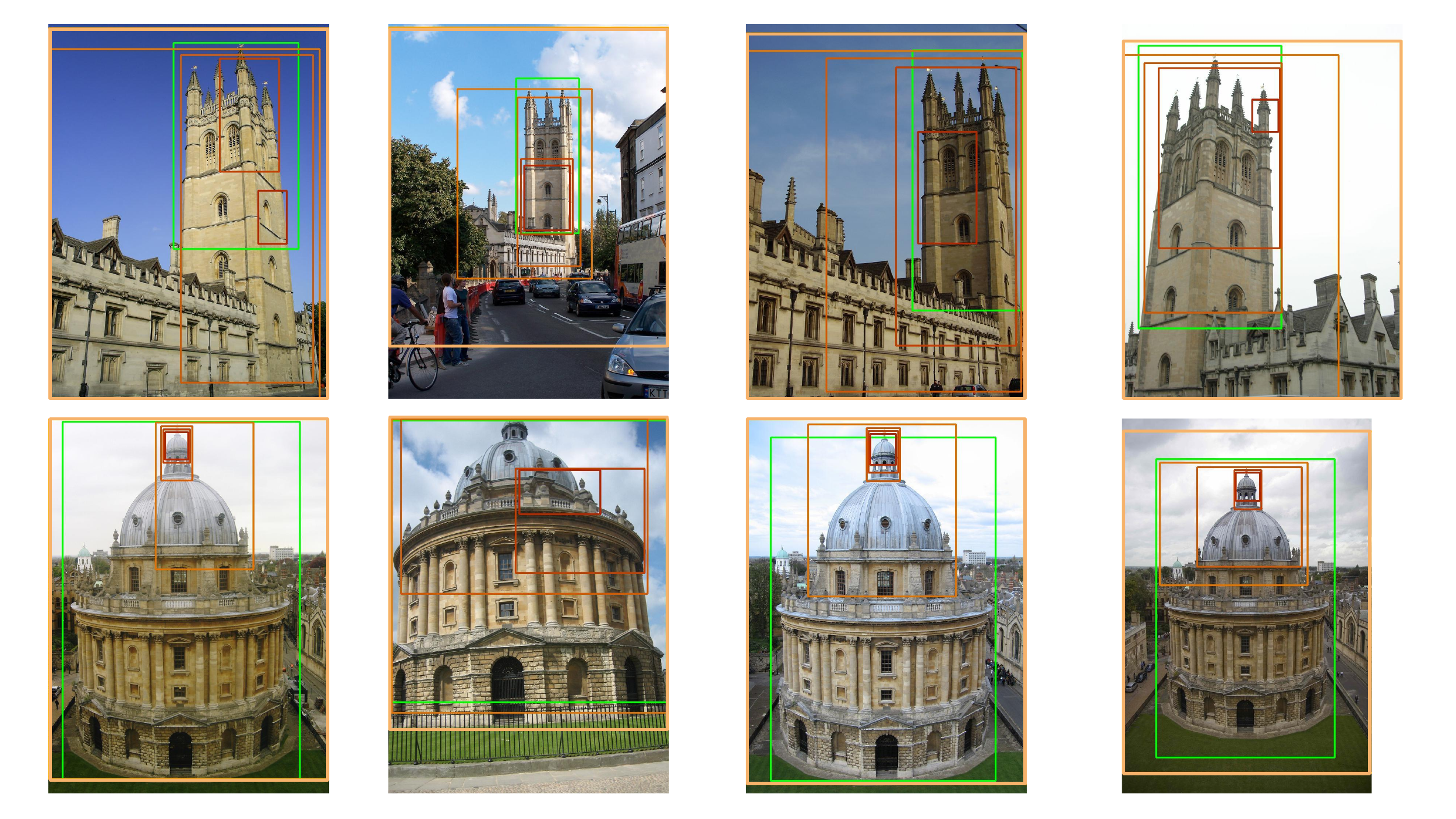}    
    \caption{Examples of regions of interest generated using CAMs. The ground truth is shown in green while the rest of colors refer to bounding boxes generated using different thresholds.}\label{fig:ROI_CAMS}
\vspace{-0.25cm}
\end{figure}

\textbf{Feature Weighting and Pooling:} In this stage, a compact representation is obtained by weighting and pooling the features. For a given class \textit{c}, we weight its features spatially, multiplying element-wise by the corresponding normalized CAM. Then, we use sum-pooling to reduce each convolutional feature map to a single value producing a K-dimensional feature vector. In our approach, our goal is to cover the extension of the objects rather than their most discriminative parts. Therefore, we consider sum-pooling instead of max-pooling. In addition, as also noted in~\cite{babenko2015aggregating,kalantidis2015cross}, sum-pooling aggregation improves performance when PCA and whitening is applied. Finally, we include the channel weighting proposed in CroW~\cite{kalantidis2015cross} to reduce channel redundancies and augment the contribution of rare features. More precisely, we first compute the proportion of non zero responses for each channel with respect to the feature map area \textit{Q\textsubscript{k}} as 
\begin{equation}
\label{eq:non_zero_responses}
Q_{k}=\frac{\sum_{i,j}^{} 1~[\chi_{i,j}^{(k)} > 0]}{WH}.
\end{equation}
Then, the channel weighting $CW_{k}$ is computed as the logarithm of the inverse channel sparsity~\cite{kalantidis2015cross}: 
\begin{equation}
\label{eq:channelweight}
CW_{k} = \log(\frac{\sum_{n=1}^K(Q_{n})}{Q_{k}}).
\end{equation}
Finally, the fixed length class vector $F^c = [f_{1}^c,f_{2}^c, ..., f_{K}^c]$ is computed as follows, 
\begin{equation}
\label{eq:weightedfeature}
f_{k}^{(c)} = CW_{k} \sum_{i=1}^W\sum_{j=1}^H \chi_{i,j}^{(k)}~CAM_{i,j}^{(c)}.
\end{equation}


\textbf{Descriptor Aggregation:} In this final stage, a descriptor \textit{D\textsubscript{I}} for each image \textit{I} is obtained by aggregating \textit{N\textsubscript{C}} class vectors. In particular, following~\cite{kalantidis2015cross,tolias2015particular}, we perform \textit{l\textsubscript{2}} normalization, PCA-whitening~\cite{jegou2012negative} and \textit{l\textsubscript{2}} normalization. Then, we combine the class vectors into a single one by summing and normalizing them. 

The remaining is selecting the classes to aggregate the descriptors. In our case, we are transferring a pre-trained network into other datasets. Therefore, we define the following two approaches:
\begin{itemize}
\item \textbf{Online Aggregation (OnA):} The top \textit{N\textsubscript{C}} predicted classes of the query image are obtained at search time (online) and the same set of classes is used to aggregate the features of each image in the dataset. This strategy generates descriptors adapted to the query. However, it has two main problems limiting its scalability: First, it requires extracting and storing CAMs for all the classes of every image from the target dataset, with the corresponding requirements in terms of computation and storage. Second, the aggregation of weighted feature maps must also be computed at query time, which slows down the retrieval process.
\item \textbf{Offline Aggregation (OfA):} The top \textit{N\textsubscript{C}} semantic classes are also predicted individually for each image in the dataset at indexing time. This is done offline and thus, no intermediate information needs to be stored, just the final descriptor. As a result, this process is more scalable than the online approach. 
\end{itemize}
\section{Experiments}\label{sec:experiments}
\begin{table}[!t]
\centering
\begin{tabular}{cc}
{\small
\begin{tabular}{@{}llcc@{}}
\toprule
    &Method& Oxford5k & Paris6k \\ \midrule
    \multirow{4}{*}{\rotatebox{90}{Baselines}}&Raw Features              & 0.396 & 0.526   \\
&Raw + Crow      & 0.420 & 0.549    \\
&Raw Features + PCA        & 0.589 & 0.662   \\
&Raw + Crow + PCA & 0.607 & 0.685   \\ \bottomrule
	\toprule
    \multirow{6}{*}{\rotatebox{90}{Network}}&
VGG-16 (Raw)              & 0.396 & 0.526   \\
&VGG-16 (64CAMs)              & 0.712 & 0.805   \\
&Resnet-50 (Raw)              & 0.389 & 0.508   \\
&Resnet-50 (64CAMs)              & 0.699 & 0.804   \\
&Densenet-161 (Raw)              & 0.339 & 0.495   \\
&Densenet-161 (64CAMs)              & 0.695 & 0.799  \\
\bottomrule
\end{tabular}}&
\hspace{-0.2cm}
{\small
\begin{tabular}{lcc}
\toprule
Aggregation & Time (s) & mAP   \\ \midrule
Raw + PCA                                  & 0.5     & 0.420 \\
1 CAM                                      & 0.5      & 0.667 \\
8 CAMs                                     & 0.6      & 0.709 \\
32 CAMs                                    & 0.9      & 0.711 \\
64 CAMs    & 1.5      & 0.712 \\ \bottomrule
\end{tabular}}\\
\vspace{0.05cm}(a)&\vspace{0.05cm}(b)\\
\end{tabular}
\vspace{-0.1cm}
\caption{a) Mean average precision comparison on Oxford5k and Paris6k for baseline methods not including CAM weighting and several network architectures used to extract CAMs. b) Actual computational cost added by using the proposed CAM weighting scheme.}
\label{tb:baselines}
\end{table}

\subsection{Datasets and Experimental Setup} 
\label{ssec:datasets}
We conduct experiments on Oxford5k Buildings~\cite{philbin2007object} and Paris6k Buildings~\cite{philbin2008lost}. Both datasets contain 55 query images to perform the search, each image annotated with a region of interest. We also consider Oxford105k and Paris106k datasets to test instance-level retrieval on a large-scale scenario. These two datasets extend Oxford5k and Paris6k with 100k distractor images collected from Flickr~\cite{philbin2007object}. Images are resized to have a minimum dimension of 720, maintaining the aspect ratio of the original image. We follow the evaluation protocol using the convolutional features of the query's annotated region of interest. We compute the PCA parameters in Paris6k when we test in Oxford5k, and vice versa. We choose the cosine similarity metric to compute the scores for each image and generate the ranked list. Finally, we use mean Average Precision (mAP) to compute the accuracy of each method.
\subsection{Network Architecture} 
\label{ssec:architecture}
In this section, we explore the use of CAMs obtained using different network architectures such as DenseNet-161~\cite{huang2016densely}, ResNet-50~\cite{he2016deep}, DecomposeMe ~\cite{alvarez2016decomposeme} and VGG-16~\cite{zhou2015cnnlocalization}. Figure~\ref{fig:qualitative_cams} shows representative CAM results for these architectures and, in Table~\ref{tb:baselines}.a we summarize the accuracy for each model. As shown in Figure~\ref{fig:qualitative_cams}, VGG-16 tends to focus on particular objects or discriminative parts of these objects rather than in the global context of the image. In addition, the length of the descriptor is 512 (compared to 2048 in ResNet-50). In addition, VGG-16 outperforms the other architectures. Therefore, we based our model in VGG-16 pre-trained on the ILSVRC ImageNet dataset~\cite{russakovsky2015imagenet} for the rest of experiments. Using this model, we extract features from the last convolutional layers (\textit{conv5\_1}, \textit{conv5\_2}, \textit{conv5\_3}) and empirically determine that \textit{conv5\_1} is the one giving the best performance. As mentioned in~\cite{zhou2015cnnlocalization}, the CAM-modified model performs worse than the original VGG-16 in the task of classification, and we verify using a simple feature aggregation that the convolutional activations are worse for the retrieval case too. For Oxford5k dataset the relative differences are of 14.8\% and 15.1\% when performing max-pooling and sum-pooling, respectively. 



\subsection{Ablation Studies} \label{ss:ablation}
The model presented in Section \ref{ss:encoding} requires two different parameters to tune: the number of class vectors aggregated \textit{N\textsubscript{C}}, and the number of classes used to build the PCA matrix, \textit{N\textsubscript{PCA}}. 
The input matrix to compute it has dimensions $N_{Im}N_{pca}\times{K}$ where $N_{Im}$ and $K$ are the number of images in the dataset and the number of feature maps of the convolutional layer considered, respectively.  

\begin{figure}[!t]
	\centering
    \hspace{-0.35cm}
	\includegraphics[scale=0.52]{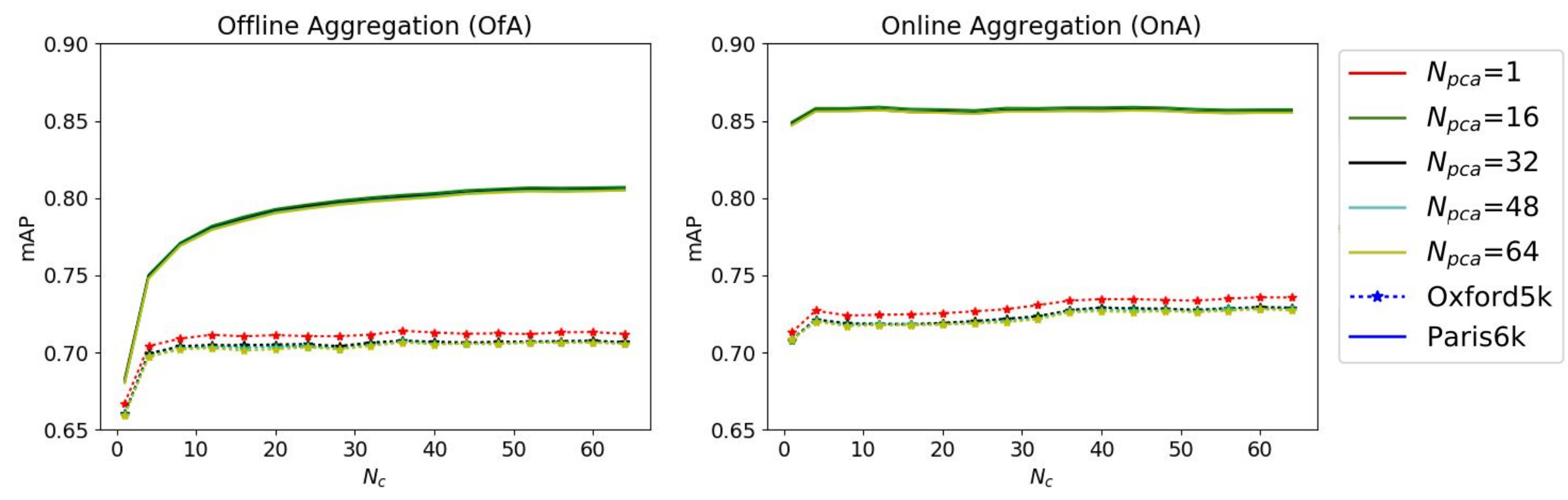}    
    \vspace{-0.75cm}
    \caption{Sensitivity of our descriptor as a function of \textit{N\textsubscript{C}} for different number of classes used to compute PCA, $N_{pca}$, for the two aggregation strategies: Online (left) and Offline (right). Straight and Dashed lines corresponds to Paris6k and Oxford5k dataset respectively.}
    \label{fig:CAMscomparison}
    \vspace{-0.3cm}
\end{figure}
\begin{figure}[!t]
	\centering
    \begin{tabular}{cc}
    \hspace{-0.3cm}\includegraphics[scale=0.415]{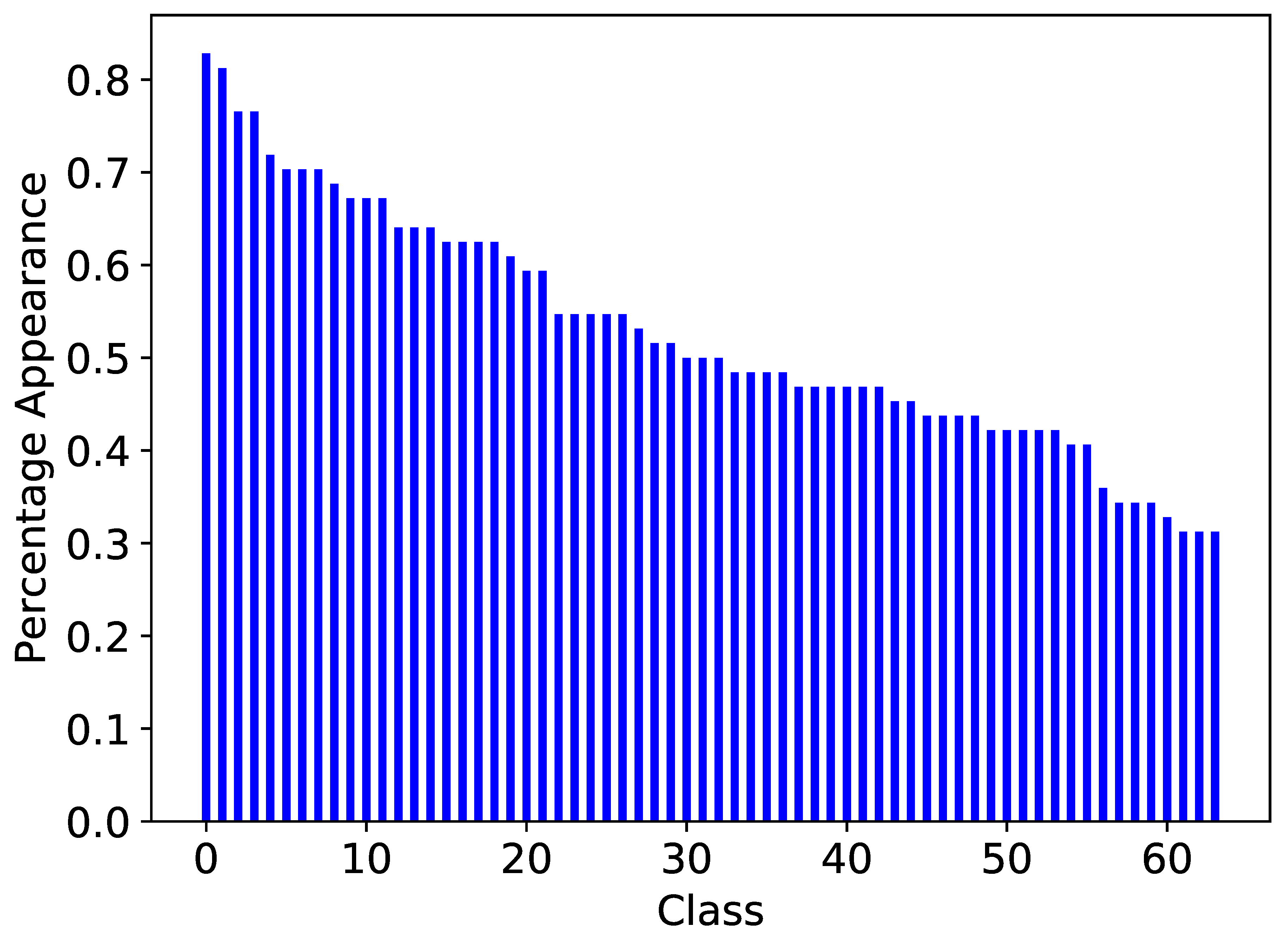}&
    \hspace{-0.3cm}\includegraphics[scale=0.43]{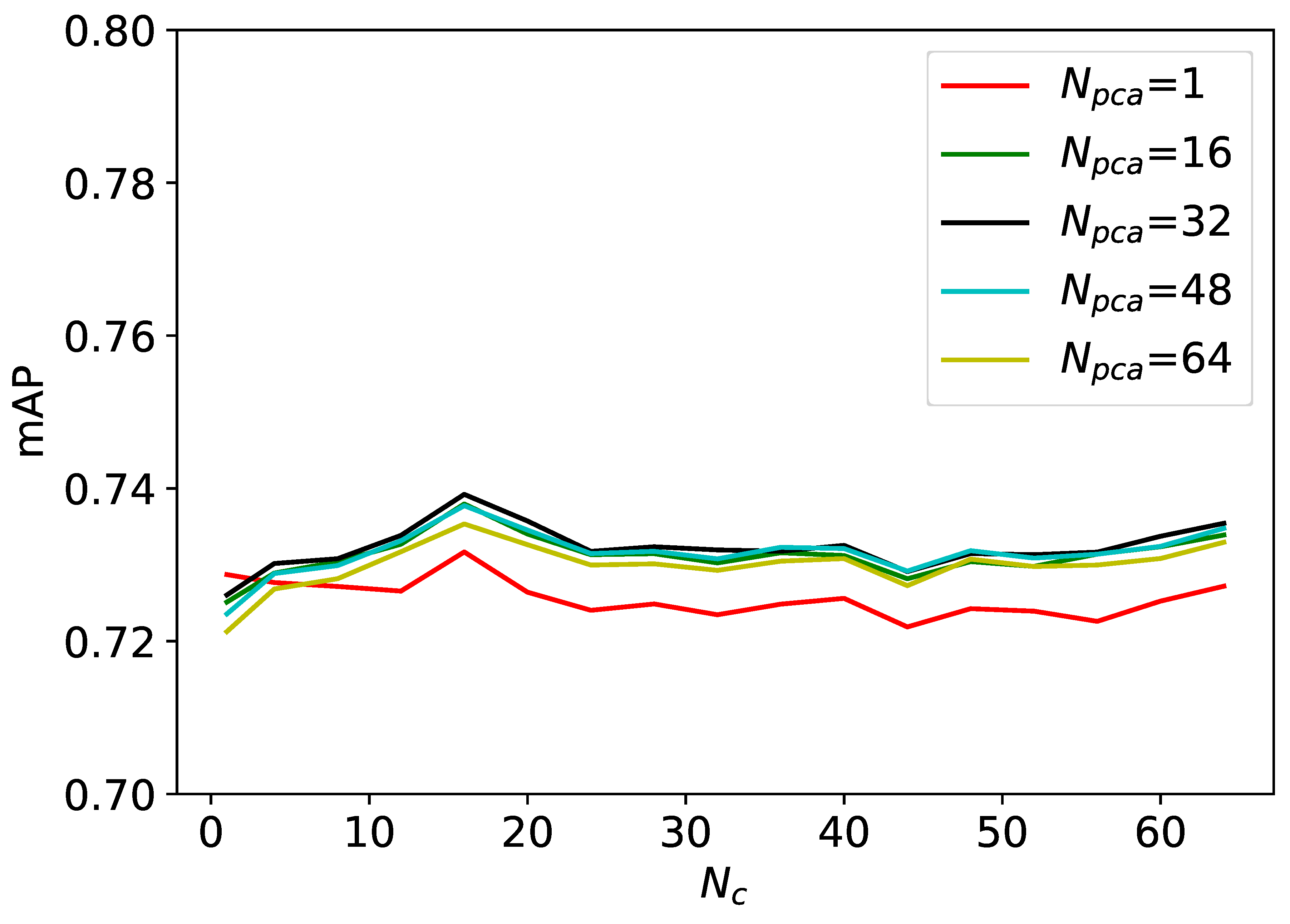}\\    
    \vspace{-0.10cm}(a)&(b)
    \end{tabular} 
    \vspace{0.05cm}
    \caption{{a) Appearance ratio of the selected classes for the 55 queries in Paris6k. b) Performance sensitivity as a function of \textit{N\textsubscript{C}} for different values of $N_{pca}$ for our Offline aggregation strategy with a set of predefined classes. Interestingly, most selected classes are related to landmarks (buildings). For instance, the first 16 classes correspond to: \textit{vault, bell cote, bannister, analog clock, movie theater, coil, pier, dome, pedestal, flagpole, church, chime, suspension bridge, birdhouse, sundial and triumphal arch}}}
   	\label{fig:Predetcomparison}
    \vspace{-0.35cm}
\end{figure}

The \textit{Online (OnA)} and \textit{Offline (OfA) Aggregations} are compared in Figure~\ref{fig:CAMscomparison} in terms of mAP as a function of the amount of top \textit{N\textsubscript{C}} classes and $N_{pca}$ classes used to compute the PCA. As a reference, the baseline mAP values obtained just sum-pooling the features, applying channel weighting and PCA can be observed in Table~\ref{tb:baselines}.a. Our technique improves that baseline without adding a large computational overhead as can also be seen in Table~\ref{tb:baselines}.b.


For the offline aggregation, the optimal \textit{N\textsubscript{C}} seems to be dataset dependent, Paris6k benefits from having more classes aggregated while the performance on Oxford5k dataset remains constant despite the number of classes. 
However, the patterns of online aggregation show that aggregating few classes (< 10) we are able to obtain a good performance for both datasets. Increasing the number of classes is also resulting in little benefit, mostly in Oxford5k dataset. It can be observed that knowing the which content is relevant and building the descriptors results accordingly in a reduction of the class vectors required, as well as a performance boost. We observe that increasing the $N_{pca}$ value does not improve the performance, suggesting that the randomness of the classes (of the target dataset) is not adding valuable information. 

To improve the performance of the offline aggregation without the practical limitations of aggregating online, we suggest restricting the total number of classes used to the most probable classes of the dataset's theme. As we have two similar building datasets, Oxford5k and Paris6k, we compute the most representative classes of the 55 Paris6k queries and use that predefined list of classes ordered by probability of appearance to obtain the image representations in Oxford5k. The results can be observed in Figure~\ref{fig:Predetcomparison}. Firstly, we see that now we are learning a better PCA transformation when increasing $N_{pca}$. As we use the same classes per every image, PCA is finding a better representation space. Secondly, we see that the mAP improves for both OfA, as now we do not have the mismatching of classes, and OnA, because the PCA is providing a better transformation.  

\subsection{Comparison to State-of-the-art Methods}\label{ss:state1}
Table~\ref{tb:qexp_rer}.a summarizes the performance of our proposal and other state-of-the-art works, all of them using an off-the-shelf VGG-16 network for image retrieval on the Oxford5k and Paris6k datasets. These results are given for a $N_{pca}$ of 1 and \textit{N\textsubscript{C}} of 64 for both approaches. 


In Paris6k benchmark, we achieve the best result with our OnA strategy, with a significant difference compared to OfA. This reflects the importance of selecting the relevant image content. We can also observe that our OfA method scales well, reaching the top performance in Oxford105k and falling behind RMAC~\cite{tolias2015particular} in Paris106k. If we are working in a particular application where we need to retrieve only specific content (e.g.~buildings), the OfA strategy could be further enhanced by doing a filtering in the pool of possible classes as described in Section~\ref{ss:ablation}. In Oxford5k benchmark, Razavian et al.~\cite{razavian2014baseline} achieve the highest performance by applying a extensive spatial search at different scales for all images in the database. However, the cost of their feature extraction is significantly higher than ours since they feed 32 image crops of resolution $576\times576$ to the CNN. In this same dataset, our OnA strategy provides the third best result using a more compact descriptor that the other techniques.

\begin{table}[!t]
\centering
{\small
\begin{tabular}{lccccc}
\toprule
\textbf{Method} & \textbf{Dim} & \textbf{Oxford5k} & \textbf{Paris6k} & \textbf{Oxford105k} & \textbf{Paris106k} \\ \midrule
SPoC~\cite{babenko2015aggregating} & 256 & 0.531 & - & 0.501 & - \\
uCroW~\cite{kalantidis2015cross} & 256 & 0.666 & 0.767 & 0.629 & 0,695 \\
CroW~\cite{kalantidis2015cross} & 512 & 0.682 & 0.796 & 0.632 & 0.710 \\
R-MAC~\cite{tolias2015particular} & 512 & 0.669 & 0.830 & 0.616 & \textbf{0.757} \\
BoW~\cite{mohedano2016bags} & 25k & 0.738 & 0.820 & 0.593 & 0.648 \\
Razavian \cite{razavian2014baseline} & 32k & \textbf{0.843} & 0.853 & - & - \\
\textbf{Ours(OnA)} & 512 & 0.736 & \textbf{0.855} & - & - \\
\textbf{Ours(OfA)} & 512 & 0.712 & 0.805 & \textbf{0.672} & 0.733 \\ \bottomrule
\end{tabular}}\\\vspace{0.05cm}(a)\\
\vspace{0.1cm}
{\small
\begin{tabular}{@{}lccccccc@{}}
\toprule
\textbf{Method} & \textbf{Dim} & \textbf{R} & \textbf{QE} & \textbf{Oxford5k} & \textbf{Paris6k} & \textbf{Oxford105k} & \textbf{Paris106k}\\ \midrule
CroW~\cite{kalantidis2015cross} & 512 & - & 10 & 0.722 & 0.855 & 0.678 & 0.797  \\
\textbf{Ours(OnA)} & 512 & - & 10 & 0.760 & 0.873 & - & -\\
\textbf{Ours(OfA)} & 512 & - & 10 & 0.730 & 0.836 & 0.712 & 0.791 \\ \bottomrule
BoW~\cite{mohedano2016bags} & 25k & 100 & 10 & 0.788 & 0.848 & 0.651 & 0.641 \\
\textbf{Ours(OnA)} & 512 & 100 & 10 & 0.780 & 0.874 & - & -  \\
\textbf{Ours(OfA)} & 512 & 100 & 10 & 0.773 & 0.838 & 0.750 & 0.780 \\ \bottomrule
RMAC~\cite{tolias2015particular} & 512 & 1000 & 5 & 0.770 & \textbf{0.877} & 0.726 & \textbf{0.817} \\
\textbf{Ours(OnA)} & 512 & 1000 & 5 & \textbf{0.811} & 0.874 & - & - \\
\textbf{Ours(OfA)} & 512 & 1000 & 5 & 0.801 & 0.855 & \textbf{0.769} & 0.800 \\ \bottomrule
\end{tabular}}\\
\vspace{0.05cm}(b)\\
\vspace{-0.2cm}
\caption{a) Comparison with the state-of-the-art CNN based retrieval methods (Off-the-shelf). b) Comparison with the state-of-the-art after applying Re-Ranking (R) or/and Query Expansion (QE). Descriptor dimensions are included in the second column (Dim).}
\label{tb:qexp_rer}
\vspace{-0.2cm}
\end{table}

\subsection{Re-Ranking and Query Expansion}\label{ss:state2}
A common approach in image retrieval is to apply some  post-processing steps for refining a first fast search such as query expansion and re-ranking~\cite{tolias2015particular,kalantidis2015cross,mohedano2016bags}.

\textbf{Query Expansion}: There exist different ways to expand a visual query as  introduced in~\cite{chum2007total, chum2011total}. We choose one of the simplest and fastest ones as in~\cite{kalantidis2015cross}, by simple updating the query descriptor for the \textit{l\textsubscript{2}} normalized sum of the top ranked $QE$ descriptors.

\textbf{Local-aware Re-Ranking}: As proposed in~\cite{philbin2007object}, a first fast ranking based on the image features can be improved with a local analysis over the top-$R$  retrieved images. This re-ranking is based on a more detailed matching between the query object and the location of this object in each top-$R$ ranked images. There are multiple ways to obtain object locations. For instance, R-MAC~\cite{tolias2015particular} applies a fast spatial search with approximate max-pooling localization. BoW~\cite{mohedano2016bags} applies re-ranking using a sliding window approach with variable bounding boxes. Our approach, in contrast, localizes objects on the images using class activation maps, as explained in Section~\ref{subs:cams}. We use the most probable classes predicted from the query to generate the regions of interest in the target images, see Figure~\ref{fig:qualitative_cams}. To obtain these regions, we first define heuristically a set of thresholds based on the normalized intensity of the CAM heatmap values. More precisely, we define a set of values 1\%, 10\%, 20\%, 30\% and 40\% of the max value of the CAM and compute bounding boxes around its largest connected component. Second, we build an image descriptor for every spatial region and compare them with the query image using the cosine distance. We keep the one with the highest score. The rationale behind using more than one threshold is to cover the variability of object dimensions in different images. Empirically, we observed that using the average heatmap of the top-2 classes improves the quality of the generated region. This is probably due to the fact that most buildings are composed by more than one class. 

We provide a comparison of our re-ranking and query expansion results with relevant state of the art methods: CroW~\cite{kalantidis2015cross} applies query expansion after the initial search. BoW and R-MAC apply first a spatial re-ranking. The number of top-images considered for these techniques varies between works. For the sake of comparison, Table~\ref{tb:qexp_rer}.b summarizes our results with their same parameters for query expansion ($QE$) and re-ranking ($R$). For the initial search, we keep $N_{pca}$ of 1 and \textit{N\textsubscript{C}} of 64 for both OnA and OfA as in the previous section. For the re-ranking process, we decrease \textit{N\textsubscript{C}} to the 6 more probable classes because, after the first search, we already have a set of relevant images and we aim at a more fine-grained comparison by looking at particular regions. In addition, taking less classes reduces the computational cost. Looking at Table~\ref{tb:qexp_rer}.b, we  observe that our proposal achieves very competitive results with a simple query expansion. Adding a re-ranking stage, the performance improves mostly in Oxford5k dataset, where we obtain the top performance. In Paris6k, we can observe that re-ranking does not increase the performance because relevant images are already on the top $QE$ of the initial list.

\section{Conclusions}\label{sec:conclusions}
In this work we proposed a technique to build compact image representations focusing on their semantic content. To this end, we employed an image encoding pipeline that makes use of a pre-trained CNN and Class Activation Maps to extract discriminative regions from the image and weight its convolutional features accordingly. Our experiments demonstrated that selecting the relevant content of an image to build the image descriptor is beneficial, and contributes to increase the retrieval performance. The proposed approach establishes a new state-of-the-art compared to methods that build image representations combining off-the-shelf features using random or fixed grid regions.

{\small
\section*{Acknowledgements}
This work has been developed in the framework of projects TEC2013-43935-R and TEC2016-75976-R, funded by the Spanish Ministerio de Economia y Competitividad and the European Regional Development Fund (ERDF). The authors also thank NVIDIA for generous hardware donations.
}
%

\bibliography{references}
\end{document}